\documentclass[sigconf]{acmart}

\usepackage{colortbl}
\usepackage{multirow}

\AtBeginDocument{%
  }

\acmSubmissionID{950}

\begin{document}

\title{Cross-Modal Dual-Causal Learning for Long-Term Action Recognition}

\author{Shaowu Xu}
\orcid{0009-0009-1852-9002}
\affiliation{%
	\institution{Beijing University of Technology}
	\department{College of Computer Science}
	\city{Chaoyang Qu}
	\state{Beijing Shi}
	\country{China}
}
\email{swxu@emails.bjut.edu.cn}

\author{Xibin Jia}
\orcid{0000-0001-8799-8042}
\authornote{Corresponding author.}
\affiliation{%
	\institution{Beijing University of Technology}
	\department{College of Computer Science}
	\city{Chaoyang Qu}
	\state{Beijing Shi}
	\country{China}}
\email{jiaxibin@bjut.edu.cn}

\author{Junyu Gao}
\orcid{0000-0002-8105-5497}
\affiliation{%
	\institution{Chinese Academy of Sciences}
	\department{Institute of Automation}
	\city{Haidian Qu}
	\state{Beijing Shi}
	\country{China}}
\email{junyu.gao@nlpr.ia.ac.cn}

\author{Qianmei Sun}
\orcid{0000-0002-1928-7772}
\affiliation{%
	\institution{Capital Medical University}
	\department{Beijing Chao-yang Hospital}
	\city{Chaoyang Qu}
	\state{Beijing Shi}
	\country{China}}
\email{sunqianmei5825@126.com}

\author{Jing Chang}
\orcid{0000-0002-2708-3921}
\affiliation{%
	\institution{Capital Medical University}
	\department{Beijing Chao-yang Hospital}
	\city{Chaoyang Qu}
	\state{Beijing Shi}
	\country{China}}
\email{cj006006@126.com}

\author{Chao Fan}
\orcid{0009-0004-1379-760X}
\affiliation{%
	\institution{Beijing University of Technology}
	\department{College of Computer Science}
	\city{Chaoyang Qu}
	\state{Beijing Shi}
	\country{China}}
\email{chao.fancripac@gmail.com}

\renewcommand{\shortauthors}{Xu et al.}

\begin{abstract}
	Long-term action recognition (LTAR) is challenging due to extended temporal spans with complex atomic action correlations and visual confounders. Although vision-language models (VLMs) have shown promise, they often rely on statistical correlations instead of causal mechanisms. Moreover, existing causality-based methods address modal-specific biases but lack cross-modal causal modeling, limiting their utility in VLM-based LTAR. This paper proposes \textbf{C}ross-\textbf{M}odal \textbf{D}ual-\textbf{C}ausal \textbf{L}earning (CMDCL), which introduces a structural causal model to uncover causal relationships between videos and label texts. 
	CMDCL addresses cross-modal biases in text embeddings via textual causal intervention and removes confounders inherent in the visual modality through visual causal intervention guided by the debiased text.
	These dual-causal interventions enable robust action representations to address LTAR challenges. Experimental results on three benchmarks including Charades, Breakfast and COIN, demonstrate the effectiveness of the proposed model. Our code is available at https://github.com/xushaowu/CMDCL.
	
\end{abstract}





\maketitle

\section{Introduction}

Spatiotemporal representation learning has significantly advanced human action recognition in videos~\cite{shamil2025utility,wu2024transferring,he2024ma}. Long-term action recognition (LTAR) remains more challenging than its short-term counterpart~\cite{yu2020rhyrnn,ryoo2019assemblenet,ryoo2021tokenlearner}, requiring modeling complex correlations among atomic actions over extended temporal spans while facing increased vulnerability to visual confounders (e.g., background and attire)~\cite{zhou2021graph,zhou2023twinformer,he2024ma}. Vision-language models (VLMs)~\cite{chen2024align,li2023blip,li2022blip} have recently emerged as powerful tools, extracting robust visual features through alignment with textual category descriptors. Fine-tuning VLM visual encoders~\cite{wu2024transferring,wu2023bidirectional,mondal2023msqnet} has proven effective for action recognition. 
However, existing research~\cite{song2024learning} indicates that VLMs often rely on spurious statistical correlations (e.g., superficial patterns like frequently co-occurring objects/backgrounds) instead of invariant causal mechanisms (e.g., essential relationships like atomic action sequences or physical constraints). This fundamentally limits their ability to discover invariant latent factors.

\begin{figure*}[ht!]
	\centering
	\includegraphics[width=\textwidth]{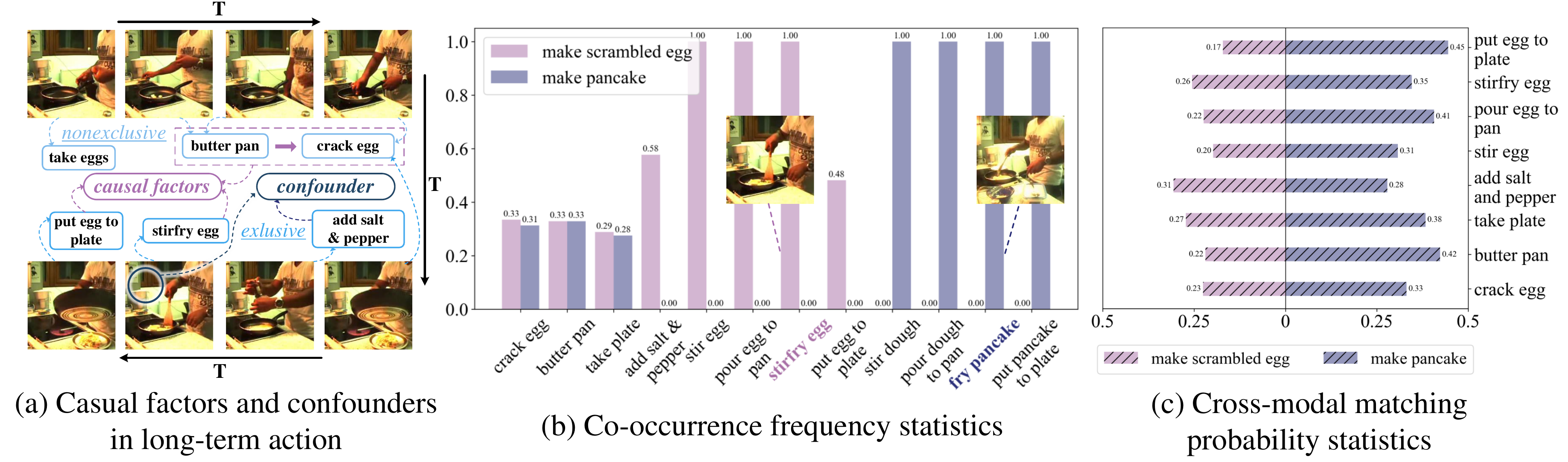} 
	\caption{Challenges in capturing true causality with VLM-based methods. 
		\textbf{(a) Causal factors and confounders in long-term actions:} Using ``make scrambled egg'' as an example, ``butter pan'' preceding ``crack egg'' illustrates sequential causality, while ``put egg to plate'' and ``stirfry egg'' demonstrate co-occurring causality. In contrast, ``add salt \& pepper'' lacks logical causality and acts as a confounder, akin to background noise. Atomic actions are further classified as exclusive or nonexclusive based on whether they appear only within specific long-term actions.  		
		\textbf{(b) Co-occurrence frequency statistics:} Analysis of the Breakfast dataset~\protect\cite{kuehne2014language} shows high co-occurrence frequencies for exclusive atomic actions within respective long-term actions (i.e., ``make scrambled egg'' and ``make pancake''), but visually similar exclusive actions across different long-term actions (e.g., ``stirfry egg'' and ``fry pancake'') pose challenges to VLMs that rely on cross-modal statistical correlations.  		
		\textbf{(c) Cross-modal matching probability statistics:} Matching probabilities between visual embeddings of atomic actions and textual embeddings of long-term actions using a pretrained VLM~\protect\cite{radford2021learning} reveal that, except for ``add salt \& pepper,'' most atomic actions in ``make scrambled egg'' correlate more with ``make pancake'' than with ``make scrambled egg.'' This highlights VLMs' failure to capture sequential and co-occurring causalities, overemphasizing visually distinctive but noncausal actions.			
	}		
	\label{fig1}
\end{figure*}

While VLMs demonstrate strong spatial vision-text alignment, their limitations become apparent in LTAR's inherent core requirements: modeling temporal atomic action correlations and suppressing visual confounders. As Figure~\ref{fig1}(a) illustrates, LTAR involves both sequential causality (e.g., ``butter pan'' preceding ``crack egg'' in ``make scrambled egg'', versus the reverse order in ``make pancake'') and co-occurring causality among atomic actions. 	
This causal diversity amplifies cross-modal bias in VLM-based models, where visually prominent but causally irrelevant atomic actions become spuriously associated with textual embeddings of semantically unrelated long-term actions. 
Consequently, as shown in Figure~\ref{fig1}(c), VLM-based models exhibit an overreliance on visually distinctive yet noncausal actions such as prioritizing ``add salt \& pepper'' for ``make scrambled egg'', which leads to suboptimal LTAR performance.
Figure~\ref{fig1}(b) further reveals co-occurring causality's vulnerability to visual confounders: exclusive atomic actions like ``stirfry egg'' are often confused with visually similar counterparts like ``fry pancake'' due to shared contextual features.

\begin{figure}[ht!]
	\centering
	\includegraphics[width=0.8\columnwidth]{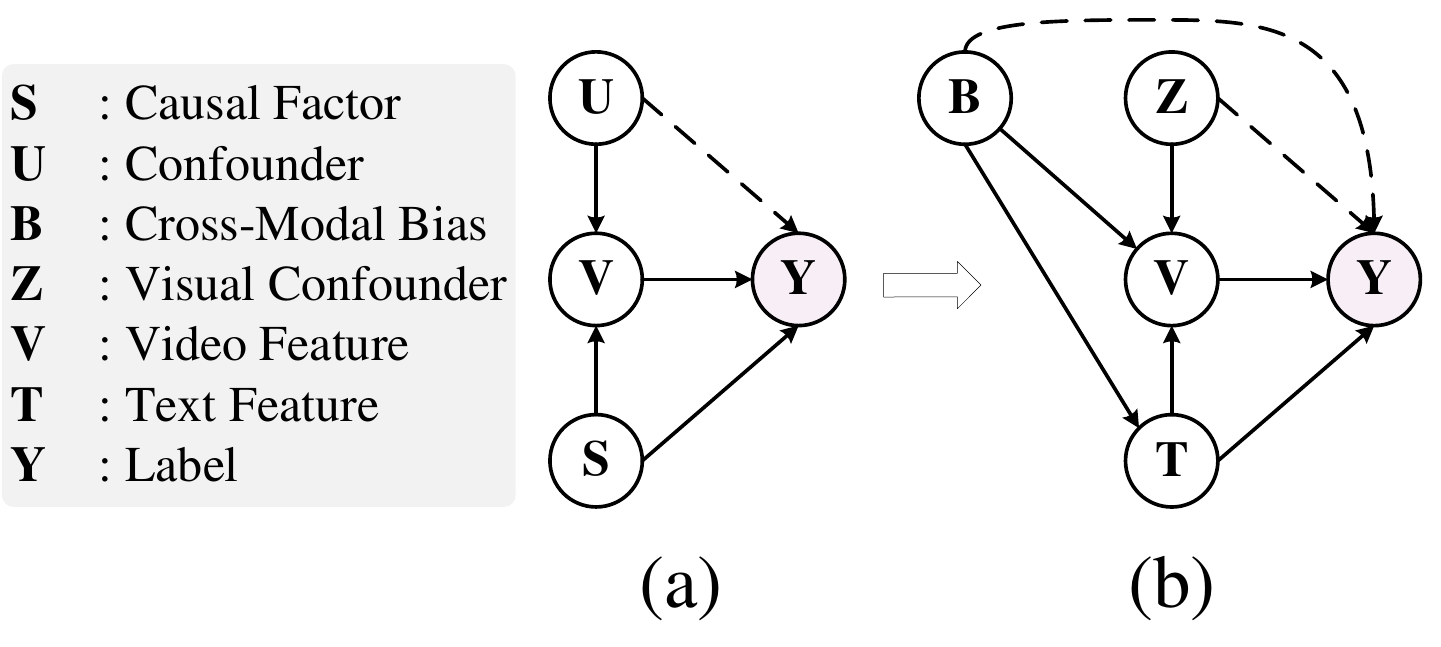} 
	\caption{The illustration of structural causal models (SCMs). (a) SCM of conventional video action recognition models. (b) SCM of pretrained VLM-based LTAR models. }
	\label{fig2}
	\vspace{-1em}
\end{figure}

Under the potential interplay between cross-modal bias and visual confounders, achieving optimal LTAR performance in VLM-based models requires Structural Causal Model (SCM)~\cite{pearl2016causal,scholkopf2022causality} to mitigate spurious statistical correlations.
However, conventional SCM, as shown in Fig.~\ref{fig2}(a), fails to characterize VLM-based LTAR. The causal factor $S$ in LTAR must encode both spatial semantics and temporal action correlations (e.g., sequential or co-occurring). When VLMs align $S$ with text features $T$, they may overemphasize spatial matching while ignoring temporal dependencies, introducing cross-modal bias $B$ that stems from spurious visual-textual correlations (e.g., frequent co-occurrence of kitchen background and cooking actions). 
This requires extending SCM to split the confounder \(U\) into cross-modal bias \(B\) and inherent visual confounders \(Z\). As shown in Figure~\ref{fig2}(b), \(B\) and \(Z\) create backdoor paths \(\{V, T\} \leftarrow B \rightarrow Y\) and \(V \leftarrow Z \rightarrow Y\), which lead the model to learn spurious correlations instead of true causal relationships.
While recent methods \cite{liu2023cross,liu2024knowledge} pay attention to recovering causal features in video tasks through causal interventions~\cite{pearl2016causal}, they remain limited for LTAR due to the lack of explicit cross-modal causal modeling and difficulty disentangling spurious correlations in untrimmed long-term videos.

To overcome these challenges, we propose \textbf{C}ross-\textbf{M}odal \textbf{D}ual-\textbf{C}ausal \textbf{L}earning (CMDCL), which learns deconfounded task-relevant correlations through causal intervention. 
First, textual causal intervention uses back-door adjustment to disentangle spurious correlations between textual and visual modalities, thereby removing cross-modal biases in text embeddings.
Second, guided by debiased text embeddings, visual causal intervention uses front-door adjustment to block back-door paths introduced by visual confounders, improving semantic consistency and causal relevance.
By integrating these dual-causal interventions, CMDCL learns robust action representations tailored for LTAR.
The main contributions of this work are summarized as follows:
\begin{itemize}
	\item We propose a novel causality-aware LTAR framework, \textbf{C}ross-\textbf{M}odal \textbf{D}ual-\textbf{C}ausal \textbf{L}earning (CMDCL), which leverages explicit causal interventions for vision-language fusion to uncover intricate causal structures and enable robust long-term action recognition.
	\item We introduce a textual causal intervention to reduce cross-modal bias by explicitly modeling the semantic correlations between action descriptions and corresponding atomic actions, eliminating spurious cross-modal correlations in pretrained VLMs.
	\item We propose a visual causal intervention that removes visual confounders through front-door adjustment guided by debiased cross-modal knowledge, while introducing an independent visual encoder to block biased information from pretrained VLMs. 
	\item Extensive experiments on the datasets, Breakfast, COIN and Charades demonstrate that CMDCL uncovers vision-language causal structures, achieving state-of-the-art performance in long-term action recognition.	
\end{itemize}

\section{Related Work}
\subsection{VLM-based Long-term Action Recognition}
The growing demand for real-world applications has propelled long-term action recognition (LTAR) research. Early works focused on modeling temporal dependencies through handcrafted architectures predefined patterns~\cite{carreira2017quo,hussein2019timeception,zhou2021graph}. The attention revolution brought fundamental shifts, with spatial-temporal transformers~\cite{zhou2023twinformer} and state-space models~\cite{islam2023efficient} enabling dynamic temporal modeling. Concurrently, vision-language models (VLMs) like CLIP~\cite{radford2021learning} and BLIP~\cite{li2022blip} demonstrated text-guided visual grounding, inspiring LTAR adaptations including ActionCLIP's~\cite{wang2021actionclip} contrastive alignment and MSQNet's~\cite{mondal2023msqnet} spatiotemporal queries. Recent innovations like BIKE~\cite{wu2023bidirectional} and Text4Vis~\cite{wu2024transferring} further bridged modalities through bidirectional knowledge transfer, whereas Video-LLaMA~\cite{zhang2023video} extended BLIP-2 with video-specific transformers. Efficiency-focused designs like MA-LMM~\cite{he2024ma} employed memory banks to handle computational complexity, marking the current state-of-the-art.

Despite these advancements, critical limitations persist in spatiotemporal causal modeling. While existing methods~\cite{wu2024transferring,zhang2023video,he2024ma} achieve strong cross-modal alignment through late fusion strategies, they fundamentally neglect two fundamental aspects of atomic action relationships: 
(1) Cross-modal bias propagation that disrupts temporal modeling through spurious visual-textual correlations between contextual environments and atomic actions; (2) Visual confounder propagation, allowing functionally distinct but visually similar actions to be misclassified due to shared environmental features in pretrained VLMs. 
These limitations originate from modeling videos as unordered frame collections rather than temporally structured processes, allowing pretrained VLM biases~\cite{song2024learning} to reinforce surface-level correlations (e.g., prioritizing incidental object co-occurrences over causal state transitions). 
Our framework addresses these gaps by employing dual causal interventions to correct cross-modal biases and visual confounders, revealing the causal relationships critical for reliable LTAR.

\vspace{-1em}
\subsection{Causal Representation Learning}

Causal mechanisms~\cite{spirtes2001causation,neuberg2003causality} underscore the limitations of relying solely on statistical dependencies—such as interpreting "add salt and pepper" as indicative of "making scrambled eggs"—when predicting counterfactual scenarios like "not adding salt and pepper still results in scrambled eggs." These mechanisms establish reasoning chains that facilitate predictions beyond observed distributions, thereby offering robust knowledge applicable to unseen contexts~\cite{scholkopf2021toward,lv2022causality}. In recent years, the integration of causality into video understanding has garnered increasing attention~\cite{niu2021counterfactual,chen2023counterfactual,liu2023counterfactual}. Causality-based methods aim to uncover invariant causal mechanisms or recover causal features to enhance video understanding~\cite{liu2023deep,wang2023deconfounding,liu2023cross,liu2024knowledge}.

However, the application of these methods to VLM-based LTAR remains underexplored. For instance, methods like~\cite{wang2023deconfounding,liu2024knowledge} address video-specific biases through causal inference but lack cross-modal causal modeling, which limits their effectiveness in VLM-based action recognition. Similarly, CMCIR~\cite{liu2023cross} employs causal reasoning for textual inference but focuses primarily on scene-object-text alignment, thereby limiting its applicability to LTAR. In this paper, we propose a cross-modal SCM tailored for VLM-based LTAR tasks. This model introduces novel dual-causal interventions designed to correct cross-modal biases and visual confounders, advancing the understanding of long-term action recognition within VLMs. We believe that the approach proposed in this paper will provide
some new ideas for subsequent work.

\begin{figure*}[ht!]
	\centering
	\includegraphics[width=\textwidth]{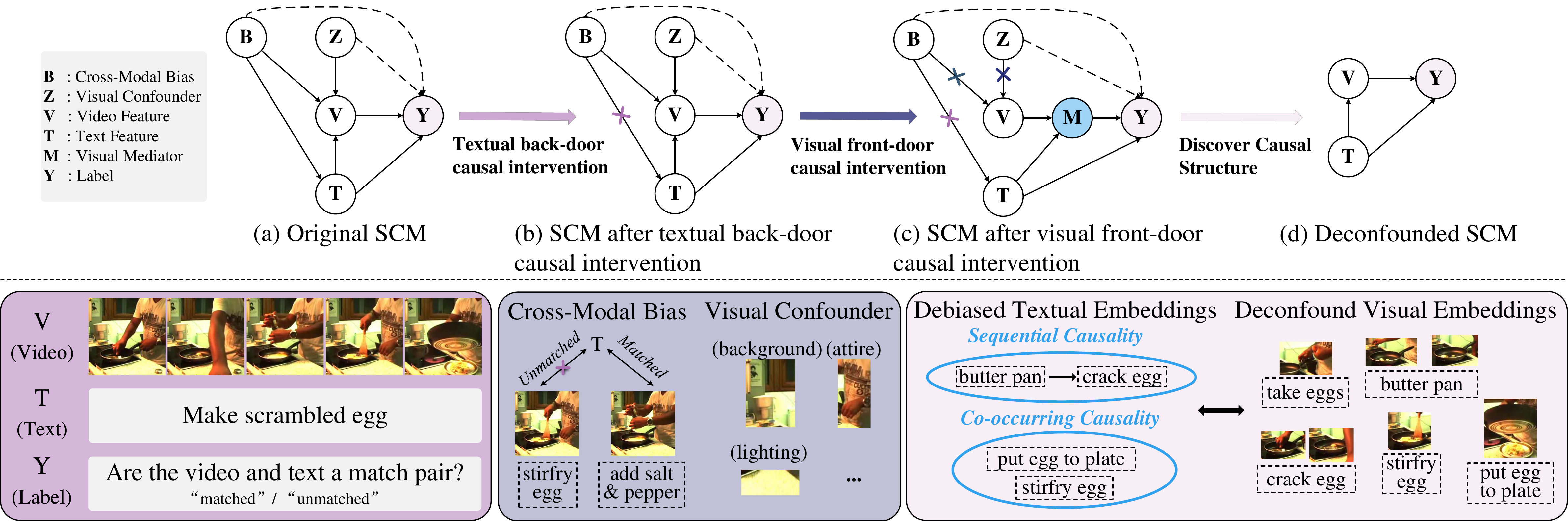} 
	\caption{The proposed causal graph of causal intervention in textual and visual modalities. The solid lines represent direct influence relation, while the dash lines represent indirect influence relation. The bottom part of the figure provides an intuitive explanation of a real LTAR sample using the proposed causal intervention.}
	\label{fig3}
\end{figure*}

\section{Method}


We formulate LTAR through a causal perspective using the SCM in Figure~\ref{fig3}. Since cross-modal biases and visual confounders are inherently unobservable, we propose the \textbf{C}ross-\textbf{M}odal \textbf{D}ual-\textbf{C}ausal \textbf{L}earning (CMDCL) framework (Figure~\ref{fig4}), to model and mitigate these factors, enabling robust representation learning for LTAR.

\subsection{Causal View of VLM-based LTAR}

For VLM-based LTAR, we employ an SCM to capture the causal effect between video-text pairs and their labels, as shown in Figure ~\ref{fig3}(a). 
Nodes represent variables, and edges denote causal relationships. 
Conventional VLM-based LTAR methods model \( V \to Y \leftarrow T \), where \( V \) represents the video, \( T \) the action text, and \( Y \) the label. This approach relies on the statistical association \( P(Y|V,T) \), often ignoring spurious correlations induced by confounders. Our framework addresses this through dual causal pathways. 
Figure ~\ref{fig3} (bottom) presents a high-level overview of cross-modal causal intervention. Below, we provide detailed interpretations of the subgraphs:

\paragraph{$T \to V \to Y$:} We learn $P(Y|V,T)$ to focus exclusively on action-relevant features in $V$ correlated with $T$. Though pretrained VLMs align $V$ and $T$ in shared space, cross-modal biases and visual confounders create backdoor interference. Our intervention aims to remove this interference to optimize LTAR performance.

\paragraph{\( T \leftarrow B \to Y \):}  
This back-door path in the textual modality arises from cross-modal bias \( B \), induced by pretrained VLMs' statistical correlations. These biases introduce spurious associations between \( V \) and \( T \), embedding noncausal atomic actions (e.g., ``add salt \& pepper'') in text features instead of true causal ones (e.g., ``stirfry egg'').  
By applying a do-operation on \( T \), we sever its direct causal dependency on \( B \), as shown in Figure~\ref{fig3}(b).

\paragraph{\( V \leftarrow \{B, Z\} \to Y \):}  
Beyond cross-modal bias $B$, visual confounders $Z$ create additional backdoor paths. By making \( V \) independent of the pretrained VLMs and blocking \( B \not\to \{V, T\} \to Y \), we ensure that \( P(Y|V,T) \) is only confounded by \( Z \). Furthermore, introducing a mediator for performing a do-operation on \( V \) blocks \( Z \not\to V \to Y \), allowing the model to learn the true causal effect \( V \to Y \leftarrow T \), as shown in Figures~\ref{fig3}(c) and (d).


The model must predict $Y$ from $V$ and $T$ while avoiding spurious correlations from $B$ or $Z$. For instance, conventional models often associate ``add salt \& pepper'' with ``make scrambled egg'' due to frequent co-occurrence, leading to misclassification of ``make fried egg'' as ``make scrambled egg'' with high confidence.
In our SCM, the non-interventional prediction is expressed using Bayes' rule:
\begin{equation}
	\resizebox{.91\linewidth}{!}{$
		P(Y|V,T) = \sum_Z \sum_B P(Y|V,T,z,b) P(z|V,T) P(b|T).
		\label{eq:non_interventional_prediction}
		$}
\end{equation}

This formulation captures both true causality $V \to Y \leftarrow T$ and spurious correlations arising from $V \leftarrow Z \to Y$ and $\{V, T\} \leftarrow B \to Y$. Applying an intervention \( do(V,T) \) can eliminate these back-door paths, removing spurious correlations and deconfounding \( \{V, T\} \) to enable true causal learning.
As cross-modal bias $B$ and visual confounder $Z$ is unobservable in VLM-based LTAR, this prevents direct computation of $P(Y|do(V,T))$ required for back-door adjustment. We therefore develop dual causal intervention modules to uncover causal structures and disentangle biases.

\subsection{Cross-Modal Dual-Causal Learning}

\begin{figure*}[ht!]
	\centering
	\includegraphics[width=0.75\textwidth]{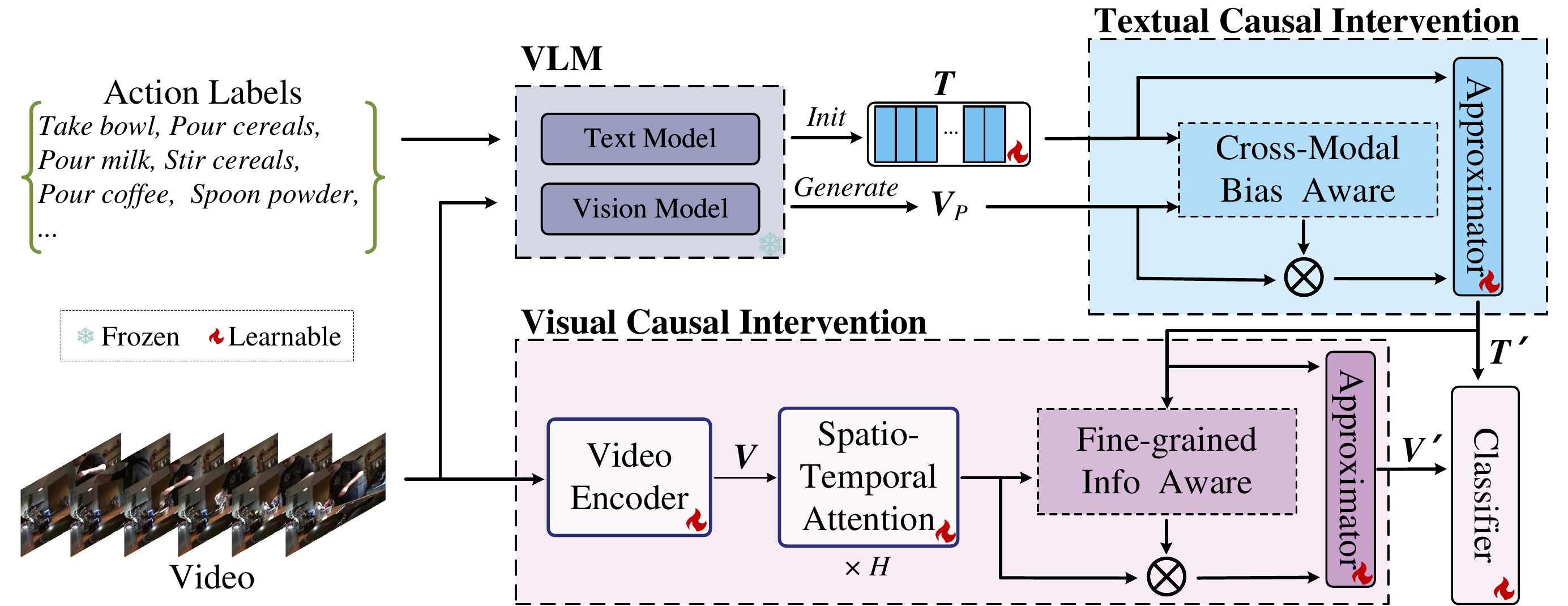} 
	\caption{The framework of CMDCL. CMDCL leverages pretrained VLMs for cross-modal knowledge extraction and employs causal interventions to mitigate cross-modal biases and visual confounders, improving LTAR performance. The Textual Causal Intervention module debiases text embeddings using back-door adjustment by calculating bias-aware scores and approximating marginal distributions of the biases. The Visual Causal Intervention module deconfounds visual embeddings via front-door adjustment, utilizing spatio-temporal attention, fine-grained information scoring, and an approximator, facilitated by a video encoder independent of VLMs that blocks visual back-door paths from VLMs. The Classifier combines debiased textual and deconfounded visual features to classify long-term actions.}
	\label{fig4}
\end{figure*}

The proposed CMDCL framework (Figure~\ref{fig4}) leverages textual and visual causal interventions to debias text embeddings, deconfound visual features, and integrate these refined features for action prediction, with detailed components described below.

\subsubsection{Textual Causal Intervention}

To address cross-modal biases in text features from pretrained VLMs, we propose the Textual Causal Intervention (TCI) module. TCI removes spurious correlations through textual back-door adjustment, targeting imbalanced co-occurrences between visual and textual features. It severs the link \( B \to T \) (Figure ~\ref{fig3}(b)) and identifies biased correlations, ensuring fair interaction with all potential cases of bias cases. The interventional distribution \( P(Y \mid V, do(T)) \) is fomulated as:
\begin{equation}
	\resizebox{.91\linewidth}{!}{$
		\begin{aligned}
			P(Y \mid V, do(T)) & = \sum_{b \in B} P(Y \mid V, do(T), b) P(b \mid V, do(T)) \\ 
			& \approx \sum_{b \in B} P(Y \mid V, do(T), b) P(b).
		\end{aligned}
		$}
	\label{eq:backdoor_adjustment}
\end{equation}

While adjustment introduces biases to identify true label causes, correlations like ``add salt \& pepper'' co-occurring with ``make scrambled egg'' can obscure matching information.
To address this, our TCI module uses visual features given by pretrained VLM as prompts (i.e. \(V_P = \{v_l^P\}_{l=1}^L \in \mathbb{R}^{L \times D}\)) to query biased features \(B\). Specifically, we estimate cross-modal bias-aware scores \(S = \{s_{c,l} \mid \text{softmax}_l\left((v_l^P)^\top t_c / \tau \right) \} \in \mathbb{R}^{C \times L}\) for learnable text embeddings \(T = \{t_c\}_{c=1}^C \in \mathbb{R}^{C \times D}\), which are initialized by pretrained VLMs. Here, \(L\) is the number of frames, \(C\) the action classes, and \(D\) the embedding dimension, \( \tau \) is the temperature parameter of the softmax.

TCI fuses scores \( S \) with \( V_P \) to obtain bias-aware embeddings \( B = \{b_c \mid \sum_{l \in L} s_{c,l} v_l^P \}  \in \mathbb{R}^{C \times D} \). 
This fused embedding serves as a surrogate for latent bias \( B \), as it captures the weighted influence of biased visual features on text feature of each action category, effectively modeling the spurious correlations introduced by \( B \).

The TCI module further employs a learnable approximator to fit the marginal distribution \( P(B) \) and applies the Normalized Weighted Geometric Mean (NWGM)~\cite{pmlrv37xuc15} to approximate debiased predictions from Eq.~\ref{eq:backdoor_adjustment}:
\begin{equation}
	\resizebox{.91\linewidth}{!}{$
		\begin{aligned}
			P(Y \mid V, do(T)) 
			& \overset{\text{NWGM}}{\approx} P(Y \mid V, \sum_B \text{concat}(do(T), b) P(b)) \\ 
			& \approx P(Y \mid V, \sum_{c=1}^C h([t_c, b_c])) = P(Y \mid V, T'),
		\end{aligned}
		$}
	\label{eq:nwgm_prediction}
\end{equation}

\noindent where \( T' = \{t_c'\}_{c=1}^C \) represents debiased text embeddings after back-door adjustment, \( [\cdot, \cdot] \) denotes concatenation, and \( h(\cdot) \) is an approximator. Iterating all samples enables integration of textual cues and cross-modal bias factors (e.g., weakening ``add salt \& pepper'' correlations with ``make scrambled egg'').

\subsubsection{Visual Causal Intervention}
We model the visual modality under the influence of cross-modal bias \( B \) and visual confounders \( Z \). In the Visual Causal Intervention (VCI) module, a video encoder independent of VLMs is introduced to extract video features \( V = \{v_l\}_{l=1}^L \in \mathbb{R}^{L \times D} \), effectively blocking \( B \to V \), while \( Z \) remains intrinsic. Notably, cross-modal bias and visual confounders operate at different scales: cross-modal bias stems from coarse-grained spurious associations between textual descriptions and atomic actions, whereas visual confounders involve fine-grained elements such as lighting and attire. These fine-grained confounders are challenging to localize in visual features, making the back-door adjustment in TCI, which relies on predefined confounders, ineffective for mitigating back-door paths caused by \( Z \).

To address this challenge, the VCI module employs a front-door adjustment strategy guided by debiased text embeddings.
Specifically, VCI assumes that TCI provides debiased text embeddings, denoted as the mediator \( M \). The suitability of \( M \) as a mediator stems from the following properties: 1) Debiased text embeddings represent intrinsic features of long-term actions, and extracting these features from visual modality is feasible, establishing \( V \to M \). 2) As \( M \) is derived from textual modality and independent of visual confounders \( Z \), there are no back-door paths between \( V \) and \( M \) involving \( Z \). 3) In VLM-based LTAR models, the classification task is effectively reframed as a matching task between visual and textual features, ensuring \( M \to Y \). 
Thus, VCI introduces the mediator \( M \), defined as \( M = T' \), forming the path \( V \to M \to Y \) to enable front-door adjustment. The interventional probability is equal to the conditional one:
\begin{equation}
		P(M=t_c' \mid do(V)) = P(M=t_c' \mid V).
	\label{eq:do_V_for_M}
\end{equation}

By intervening on \( V \), VCI can block the back-door path \( Y \leftarrow Z \to V \to M \) , even when \( Z \) is unobservable:
\begin{equation}
	\resizebox{.91\linewidth}{!}
	{$
		P(Y \mid do(M = t_c'), T) = \sum_{\hat{v}_c \in \hat{V}} P(V = \hat{v}_c) P(Y \mid V = \hat{v}_c, M = t_c', T),
		$}
	\label{eq:cut_back_door_of_V}
\end{equation}


\noindent where \( \hat{V} = \{\hat{v}_c\}_{c=1}^C \) represents the set of visually emphasized features when \(t_c'\) serves as the mediator, isolating the direct effect of visual features on the outcome, hence ensuring the causal estimate is not confounded by \(Z\). 
To derive \( \hat{V} \), VCI employs spatio-temporal attention mechanisms to enhance video representation and compute fine-grained, information-aware scores. Given \( V \), the process involves applying multi-head self-attention across \( H \) layers followed by layer normalization. The output \( V^{(H)} \) is then utilized to calculate the scores \( S^F \), with each score \( s_{c,l,d}^F \) computed using the formula \( s_{c,l,d}^F = \text{softmax}(v_{l,d}^{(H)} t_{c,d}' \, / \, \tau) \), where \( \tau \) is the temperature parameter. Each \(\hat{v}_c\) is finally defined as \( \sum_{l=1}^L s_{c,l,d}^F v_{l,d}^{(H)} \in \mathbb{R}^D \), effectively isolating key features for causal analysis.

By introducing debiased text embeddings as the mediator, VCI applies the front-door adjustment to perform the do-operation on \( V \), eliminating the interference of visual confounders on the causal estimation of \( Y \). Thus, through combining Eq.~\ref{eq:do_V_for_M} and Eq.~\ref{eq:cut_back_door_of_V}, we have: 
\begin{equation}
	\resizebox{.91\linewidth}{!}{$
		\begin{aligned}
			& P(Y \mid do(V), T) = \sum_{t_c' \in T'} P(M = t' \mid do(V)) P(Y \mid do(M = t'), T) \\
			& = \sum_{t_c' \in T'} \sum_{\hat{v}_c \in \hat{V}} P(M = t_c' \mid V) P(V=\hat{v}_c) P(Y \mid M = t_c', V=\hat{v}_c, T).
		\end{aligned}
		$}
	\label{eq:front_door_intervention}
\end{equation}

To simplify modeling, we adopt NWGM~\cite{pmlrv37xuc15}, incorporating sampling at the feature level:
\begin{equation}
	\resizebox{.91\linewidth}{!}{$
		\begin{aligned}
			P(Y \mid do(V), T) 
			\overset{\text{NWGM}}{\approx} 
			P(Y \mid \sum_{t_c' \in T'} \sum_{\hat{v}_c \in \hat{V}} \text{concat}(M = t_c', V = \hat{v}_c,) \\
			\phantom{P(} \times P(M = t_c' \mid V) P(V = \hat{v}_c), T).
		\end{aligned}
		$}
	\label{eq:nwgm_vci}
\end{equation}

Finally, VCI employs a learnable approximator \( g(\cdot) \) to approximate \( P(M = t_c' \mid V) \) and \( P(V = \hat{v}_c) \). This produces deconfounded visual embeddings \( V' = \{v_c'\}_{c=1}^C \), enabling:
\begin{equation}
	\resizebox{.91\linewidth}{!}{$
		P(Y \mid do(V), T) 
		\approx P(Y \mid \sum_{c=1}^C g([t_c', \hat{v}_c]), V) = P(Y \mid V', T).
		\label{eq:final_vci}
		$}
\end{equation}

\subsubsection{Classifier}
After obtaining deconfounded visual embeddings \( V' \) and debiased text embeddings \( T' \), we feed them into the classifier module for long-term action recognition. The predicted probability for class \(c\) is defined as:
\begin{equation}
	\resizebox{.91\linewidth}{!}{$
		P(Y = c \mid \text{do}(V, T)) = \text{Softmax}\Bigl(\sum_{d=1}^D w_{c,d} \cdot f(T', V')_{c,d} + b_c\Bigr),
		\label{eq:cmi_prediction}
		$}
\end{equation}
\noindent where \( w_{c,d} \) and \( b_c \) are the learnable parameters for class \(c\), and \( f(\cdot, \cdot) \) represents the cross-modal interaction function.

\begin{table*}[ht!]
	\centering
	\caption{
		Comparisons of Acc (\%) and mAP (\%) with SOTA methods on Breakfast and COIN. Breakfast evaluates top-1 Acc (Acc@1) while COIN reports both Acc@1 and top-5 Acc (Acc@5).
	}
	\label{tab:combined}
	\renewcommand{\arraystretch}{0.8}
	\setlength{\extrarowheight}{0pt}
	\addtolength{\extrarowheight}{\aboverulesep}
	\addtolength{\extrarowheight}{\belowrulesep}
	\setlength{\belowrulesep}{0pt}
		\begin{tabular}{lrrrrrrr} 
			\toprule
			\multirow{2}{*}{Method} & \multirow{2}{*}{Venue} & \multirow{2}{*}{Pre-training} & \multicolumn{2}{c}{Breakfast} & \multicolumn{3}{c}{COIN} \\
			\cmidrule(lr){4-5} \cmidrule(lr){6-8}
			&                        &                                & Acc@1           & mAP           & Acc@1          & Acc@5        & mAP           \\ 
			\hline\hline
			\multicolumn{8}{l}{{\cellcolor[rgb]{0.961,0.961,0.961}}\textit{Methods without pretrained VLMs}} \\ 
			\midrule
			I3D~\cite{carreira2017quo}           & CVPR'17      & -                & 58.61           & 47.05           & -              & -            & -              \\
			ActionVlad~\cite{girdhar2017actionvlad} & CVPR'19   & K400             & 65.48           & 60.20           & -              & -            & -              \\
			VideoGraph~\cite{hussein2019videograph} & arXiv'19  & -                & 69.45           & 63.14           & -              & -            & -              \\
			Timeception~\cite{hussein2019timeception} & CVPR'19 & -                & 71.30           & 61.82           & -              & -            & 57.99          \\
			GHRM~\cite{zhou2021graph}           & CVPR'21      & K400             & 75.49           & 65.86           & -              & -            & 64.63          \\
			D-Sprv~\cite{lin2022learning}       & CVPR'22      & HowTo100M        & 89.90           & -               & 90.00          & -            & -              \\
			TwinFormer~\cite{zhou2023twinformer}& TMM'23       & K400             & 79.16           & 67.64           & -              & -            & \underline{67.81} \\
			CMCIR~\cite{liu2023cross}           & TPAMI'23     & IN-22K+K600      & 89.01           & -               & 89.02          & 92.53        & -              \\ 
			TranS4mer~\cite{islam2023efficient} & CVPR’23      & K600             & 90.27           & -               & 89.23          & -            & -              \\
			S5~\cite{wang2023selective}         & CVPR'23      & K600             & 90.70           & -               & 90.80          & -            & -              \\ 
			VideoMamba~\cite{li2024videomamba}  & ECCV'24      & K400             & 94.30           & -               & 86.20          & -            & -              \\ 
			KCMM~\cite{islam2022long}           & PR'25        & K400             & 93.45           & -               & 88.11          & -            & -              \\			
			\hline\hline
			\multicolumn{8}{l}{{\cellcolor[rgb]{0.961,0.961,0.961}}\textit{Methods with pretrained VLMs}} \\ 
			\midrule
			BIKE~\cite{wu2023bidirectional}    & CVPR'23      & WIT-400M+K400     & \underline{95.77}& 59.96           & 90.62          & 99.24        & 65.15          \\
			Text4Vis~\cite{wu2024transferring} & IJCV'24      & WIT-400M+K400     & 95.49            & 60.49           & 91.98          & \underline{99.27}  & 63.39          \\
			\multirow{3}{*}{MA-LMM~\cite{he2024ma}} 
			& \multirow{3}{*}{CVPR'24} 
			& LAION-400M+   & \multirow{3}{*}{93.00} & \multirow{3}{*}{\underline{71.84}} & \multirow{3}{*}{\textbf{93.20}} & \multirow{3}{*}{96.90} & \multirow{3}{*}{\underline{67.67}} \\
			&               & ShareGPT-70K+   &                    &                 &                  &              &                   \\
			&               & merged-30M      &                    &                 &                  &              &                   \\ 
			\hline
			CMDCL (Ours)                      &               & WIT-400M+K400     & \textbf{96.62}   & \textbf{89.10}  & \underline{92.28}& \textbf{99.28} & \textbf{85.51} \\ 
			\bottomrule
		\end{tabular}
\end{table*}

\section{Experiments}

\subsection{Datasets and Setups}
We evaluate on three established benchmarks.
The \textbf{Breakfast} dataset~\cite{kuehne2014language} comprises 1712 cooking videos averaging 2.3 minutes, featuring 10 activities performed by 52 actors across multiple kitchens. Videos are labeled with 48 atomic actions, averaging 6 per video. Following~\cite{he2024ma}, the dataset is split into 1357 training and 355 testing videos.
The \textbf{COIN} dataset~\cite{tang2019coin} contains 11827 YouTube videos spanning 180 tasks across 12 domains. Each 2.36-minute video consists of 3.91 atomic actions on average, drawn from 778 atomic actions. Following~\cite{tang2019coin}, the dataset is split into 9030 training and 2797 testing videos.
The \textbf{Charades} dataset~\cite{sigurdsson2016hollywood} includes 9848 videos of everyday activities by 267 individuals. Each 30-second video is annotated with an average of 6.8 atomic actions from 157 atomic action classes. Following~\cite{wu2024transferring}, we use 7985 videos for training and 1863 for testing. 


Our implementation employs CLIP~\cite{radford2021learning} as the VLM backbone, TimeSformer~\cite{bertasius2021space} for video encoding, and Q-Former~\cite{li2023blip} for cross-modal classification. Models are trained for 80 epochs with a cosine decay scheduler and an initial learning rate of \(1\mathrm{e}^{-5}\) using the Adam optimizer. Given the multi-atomic nature of LTAR, mean average precision (mAP) provides critical evaluation by capturing concurrent spatiotemporal dependencies among multiple atomic actions, while accuracy (Acc) measures single-label performance. Following~\cite{zhou2023twinformer,wu2024transferring}, we maintain separate models per task configuration. All experiments are conducted on an RTX 3090 GPU.

\subsection{Comparative Experiments}
\noindent \textbf{Breakfast.} The Breakfast dataset supports dual evaluation. Table~\ref{tab:combined} presents that VLM-based methods (e.g., Text4Vis~\cite{wu2024transferring}, BIKE~\cite{wu2023bidirectional}) achieve higher Acc but only a comparable mAP to non-VLM methods (e.g., GHRM~\cite{zhou2021graph}, TwinFormer~\cite{zhou2023twinformer}), indicating that they excel at recognizing atomic actions that occur exclusively in specific long-term actions but struggle to model causal relationships among all atomic actions. Though CMCIR~\cite{CMCIR} adopts causal modeling, its lack of cross-modal intervention limits performance relative to VLM-based methods.
CMDCL uniquely integrates pretrained VLMs and dual causal intervention, addressing cross-modal bias and visual confounders for superior performance.

\noindent \textbf{COIN.} 
Unlike Breakfast which uses Acc@1 and mAP, COIN extends evaluation with Acc@5 given its richer long-term action categories. As Table~\ref{tab:combined} shows, CMDCL achieves state-of-the-art performance except for a slightly lower Acc@1 than  MA-LMM~\cite{he2024ma}.
This arises from COIN's pseudo-confounders (e.g., text-overlaid video segments where visual frames contain textual summaries of content) that correlate with content but lack action causality. MA-LMM's VLM-LLM framework exploits these pseudo-confounders through enhanced statistical correlation learning, improving single-label classification but risking overfitting non-causal elements. This compromises causal modeling and multi-label generalization, thus yielding lower Acc@5 and mAP than our CMDCL.


\begin{figure*}[t!]
	\centering
	\includegraphics[width=\textwidth]{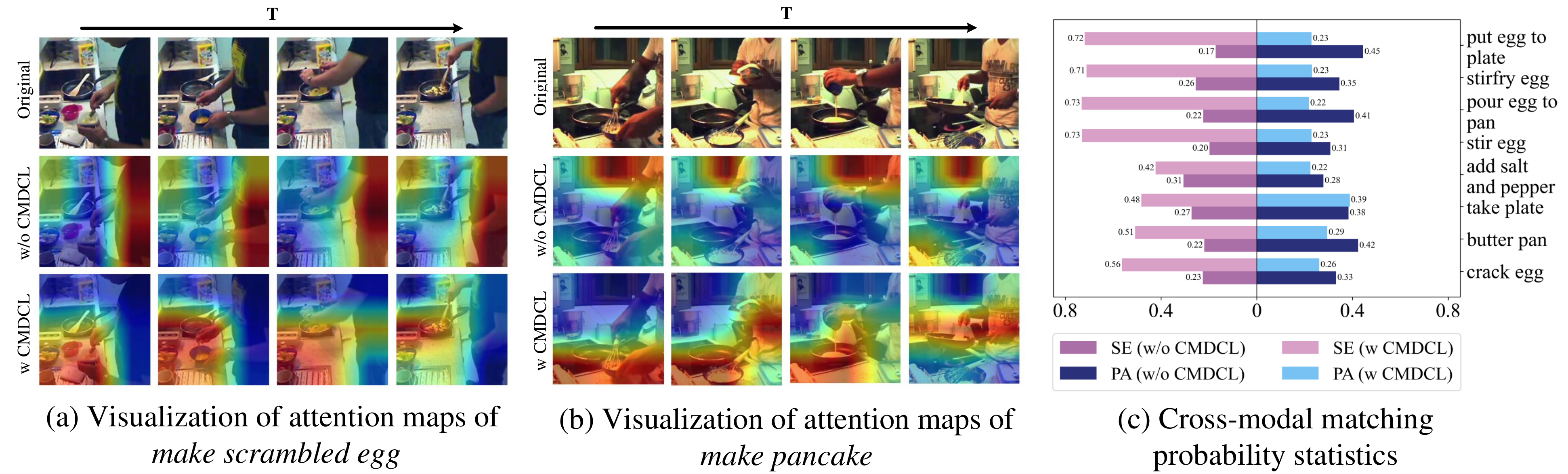} 
	\caption{Visual analysis of the deconfounding effect. \textbf{(a)} and \textbf{(b)} present attention maps for the long-term actions ``make scrambled egg'' and ``make pancake'', respectively, generated using the method from~\citep{selvaraju2017grad}. The first row displays original frame sequences, while the second and third rows show attention maps from VLM-based models without and with our CMDCL framework. Observations reveal that model without CMDCL exhibit erroneous attention focus on visual confounders (e.g., human attire and background elements), whereas our CMDCL effectively guides the model to concentrate on causality-related regions, particularly areas showing hand-object interactions with cooking utensils. 
		\textbf{(c)} provides cross-modal matching probability statistics between visual embeddings of atomic actions and textual embeddings of long-term actions (SE:make scrambled egg, PA:make pancake), following the format of Figure~\ref{fig1}(c). Our CMDCL strengthens visual-text alignment for atomic actions with co-occurring causality (e.g., ``stirfry egg''→SE), while weakening associations for noncausal actions (e.g., ``add salt \& pepper''→SE), thereby demonstrating effective causal pattern learning and cross-modal bias mitigation.
	}
	
	\label{fig7}
\end{figure*}

\noindent \textbf{Charades.} 
As the Charades dataset provides only atomic action category labels, we evaluate model performance using mAP. Table~\ref{tab:charades} shows that methods trained on pre-trimmed single-label videos (e.g., TokenLearner~\cite{ryoo2021tokenlearner}) achieve higher mAP, as pre-trimming simplifies video scenes. In contrast, CMDCL, which does not rely on pre-trimmed videos, achieves 51.1\% mAP among VLM-based methods, surpassing BIKE (trained with pre-trimmed videos) and MSQNet~\cite{mondal2023msqnet} (trained without pre-trimmed videos). This highlights CMDCL’s robust recognition capabilities in more complex long-video scenarios without requiring pre-trimmed data.

\begin{table}[ht!]
	\centering
	\caption{Comparisons on Charades. “Pre-trimmed” indicates that the method trains the model using pre-trimmed single-label videos (each video corresponds to one atomic action) instead of using complete multi-label videos.}
	\label{tab:charades}
		\renewcommand{\arraystretch}{1}
		\begin{tabular}{lcr} 
			\toprule
			Method      & Pre-trimmed & mAP(\%)                                                                 \\ 
			\hline\hline
			\multicolumn{3}{l}{{\cellcolor[rgb]{0.961,0.961,0.961}}\textit{Methods without pretrained VLMs}}    \\ 
			\midrule
			ActionVLAD \cite{girdhar2017actionvlad}  & \checkmark  & 21.0                                                                       \\
			Rhyrnn \cite{yu2020rhyrnn}      & \checkmark  & 25.4                                                                   \\
			I3D \cite{carreira2017quo}         & \checkmark  & 32.9                                                                   \\
			STM \cite{jiang2019stm}         & \checkmark  & 35.3                                                                   \\
			VideoGraph \cite{hussein2019videograph}  & \checkmark  & 37.8                                                                       \\
			MViT-B \cite{fan2021multiscale}      & \checkmark  & 43.9                                                                       \\
			AssembleNet \cite{ryoo2019assemblenet} & \checkmark  & 58.6                                                                       \\
			TokenLearner \cite{ryoo2021tokenlearner}& \checkmark  & 66.3                                                                       \\
			Timeception \cite{hussein2019timeception} & ×    & 37.2                                                                       \\
			GHRM \cite{zhou2021graph}        & ×    & 38.3                                                                   \\
			TwinFormer \cite{zhou2023twinformer}  & ×    & 43.6                                                                        \\ 
			\hline\hline
			\multicolumn{3}{l}{{\cellcolor[rgb]{0.961,0.961,0.961}}\textit{Methods with pretrained VLMs}}       \\ 
			\midrule
			ActionCLIP \cite{wang2021actionclip}    & \checkmark       & 44.3                                                                   \\
			Text4Vis \cite{wu2024transferring}    & \checkmark       & 46.0                                                                   \\
			BIKE \cite{wu2023bidirectional}        & \checkmark       & 50.4                                                                   \\
			MSQNet \cite{mondal2023msqnet}      & ×         & 47.6                                                                       \\
			\hline
			CMDCL (Ours) & ×         & \textbf{51.1}                                                                   \\
			\bottomrule
		\end{tabular}
\end{table}

\begin{figure}[t!]
	\centering
	\includegraphics[width=\columnwidth]{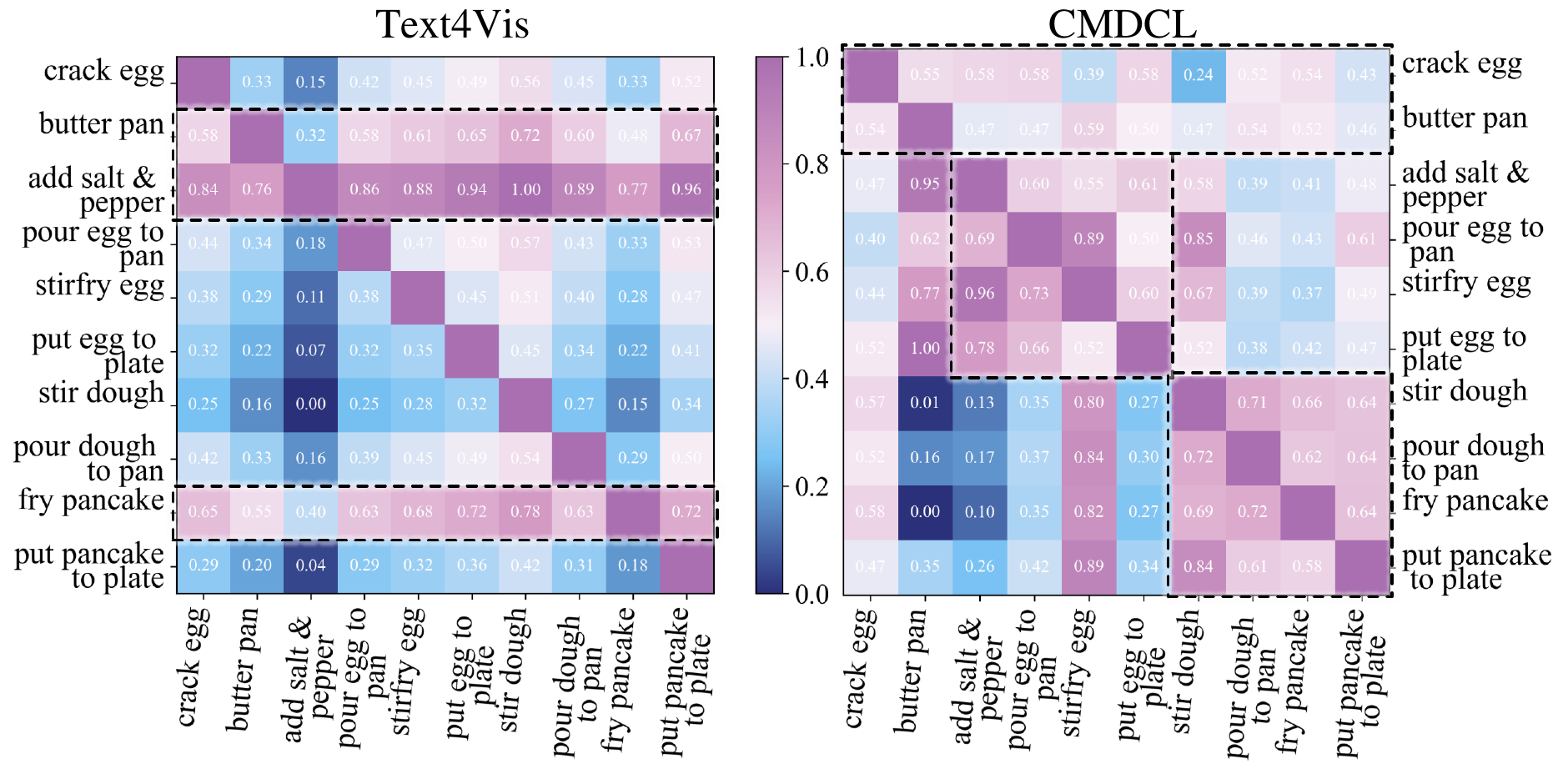} 
	\caption{Visualization of class-to-class co-classification probabilities of Text4Vis~\protect\cite{wu2024transferring} (left) and our CMDCL (right). The matrices present normalized co-classification probability for each pair of atomic actions, where the value in row \(i\) and column \(j\) indicates the probability that atomic action \(i\) is classified when atomic action \(j\) is classified. 
	}
	\vspace{-1em}
	\label{fig5}
\end{figure}

\begin{figure}[t!]
	\centering
	\includegraphics[width=\columnwidth]{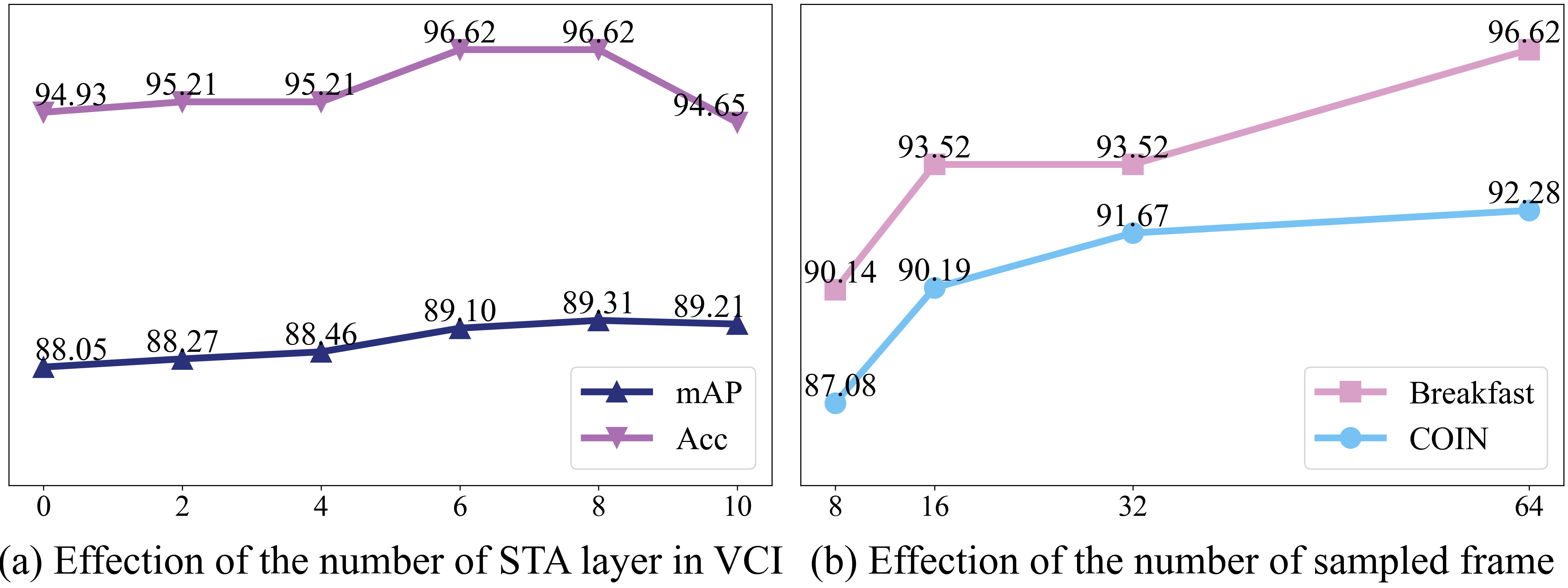} 
	\caption{Effection of hyperparameters. (a) Effection of the number of spatio-temporal attention (STA) layers in the VCI module using Acc and mAP metrics on the Breakfast dataset. (b) Effection of different numbers of sampled frames using Acc as the metric on the Breakfast and COIN datasets.}
	\label{fig6}
\end{figure}

\subsection{Analytical Experiments}

\noindent \textbf{Ablation Study.} 
Table~\ref{tab:ablation} evaluates component contributions under state-of-the-art settings. The TCI module addresses cross-modal bias, effectively preventing atomic actions from being correlated with incorrect long-term actions, yet exhibits limited constraints on frame-level discriminability. Conversely, the VCI module suppresses visual confounders through fine-grained feature rectification, but lacks explicit constraints for cross-modal matching.  This explains Variant 2 (TCI-only)'s high accuracy (95.21\%/96.62\%) but limited mAP (85.38\%/89.10\%), versus Variant 3 (VCI-only)'s superior mAP (88.31\%/89.10\%) with accuracy trade-offs (93.80\%/96.62\%). 
The full CMDCL framework's synergistic integration of both modules yields complementary advantages, achieving maximal gains in both metrics, thereby confirming the necessity of dual causal learning. 

\noindent \textbf{Visual Confounder Suppression Analysis.} 
Our CMDCL effectively suppresses visual confounders through the VCI module. As shown in Figure~\ref{fig7}(a)-(b), the baseline model exhibits attention dispersion towards semantically irrelevant regions (e.g., attire and background), while CMDCL focuses precisely on hand-object interactions through causal attention convergence. 
This improvement originates from VCI's front-door adjustment mechanism that filters out noncausal visual features (e.g., kitchen background or clothing textures) while amplifying temporally critical hand-object interaction patterns. The converged attention patterns demonstrate enhanced robustness against environmental variations, maintaining focus on causal action semantics across different culinary scenarios.
These results validates VCI's capability to enhance VCI's capacity in enhancing fine-grained feature reliability.

\noindent \textbf{Cross-modal Bias Mitigation Verification.} 
TCI's causal intervention is validated via  cross-modal matching statistics (Figure~\ref{fig7}(c)). The baseline exhibits biased matching visual embeddings of atomic actions from ``make scrambled eggs''  with the textual embeddings of ``make pancake'', while CMDCL shows precise matching of causal atomic actions like ``stirfry egg'' to their corresponding long-term actions. 
Such disparity originates from TCI's ability to disentangle utensil-context biases implicitly embedded in textual descriptors by applying back-door adjustment. As a result, it enforces a stricter alignment between the semantics of atomic actions and their corresponding causal visual patterns. These findings provide strong evidence that TCI effectively mitigates spurious cross-modal correlations through causal intervention in the textual modality.

\begin{table}[t!]
	\centering
	\caption{Impact of the TCI and VCI modules in CMDCL on the Breakfast and COIN datasets.}
	\label{tab:ablation}
	\renewcommand{\arraystretch}{0.8}
	\setlength{\extrarowheight}{0pt}
	\addtolength{\extrarowheight}{\aboverulesep}
	\addtolength{\extrarowheight}{\belowrulesep}
	\setlength{\aboverulesep}{0pt}
	\setlength{\belowrulesep}{0pt}
	\resizebox{\columnwidth}{!}{
		\begin{tabular}{lccrrrr} 
			\toprule
			\multirow{2}{*}{Method} & \multirow{2}{*}{TCI} & \multirow{2}{*}{VCI} & \multicolumn{2}{c}{Breakfast} & \multicolumn{2}{c}{COIN}  \\
			\cmidrule(lr){4-5} \cmidrule(lr){6-7}
			&                      &                      & Acc            & mAP          & Acc          & mAP         \\
			\hline\hline
			Variant 1 & ×              & ×               & 91.49     & 81.72     & 87.78     & 76.93     \\
			Variant 2 & \checkmark     & ×               & \underline{95.21}     & 85.38     & \underline{92.14}     & 80.79     \\
			Variant 3 & ×              & \checkmark      & 93.80     & \underline{88.31}     & 90.96     & \underline{82.13}     \\
			\midrule
			CMDCL & \checkmark     & \checkmark      & \textbf{96.62}     & \textbf{89.10}     & \textbf{92.28}     & \textbf{85.51}     \\ 
			\bottomrule
		\end{tabular}
	}
	\vspace{-1em}
\end{table}

\noindent \textbf{Atomic Action Discriminability Analysis.} 
Figure~\ref{fig5} visualizes atomic action co-classification probabilities comparing CMDCL with Text4Vis~\cite{wu2024transferring}. CMDCL demonstrates two critical improvements: 1) \textit{Mitigating cross-modal bias}: Unlike Text4Vis' uniform probability distribution for ``butter pan'' across actions, CMDCL strengthens sequential causality - correlating ``butter pan'' with ``stirfry egg'' and ``crack egg'' with ``fry pancake'' - revealing recognition of temporal order distinguishing scrambled eggs vs. pancakes. 2) \textit{Suppressing visual confounders}: While Text4Vis falsely links visually distinct actions (``add salt \& pepper'' vs. ``fry pancake''), CMDCL enhances co-classification among exclusive atomic actions within each long-term task (e.g., scrambled egg-specific actions), effectively capturing co-occurring causality patterns.

\noindent \textbf{Effect of hyperparameters.} 
Figure~\ref{fig6}(a) illustrates that although the configuration with 8 stacked STA layers achieves the highest mAP of 89.31\%, the improvement over the 6-layer version (89.10\%) is marginal. In addition, both configurations yield the same top-1 accuracy of 96.62\%, indicating that the additional layers contribute little to overall classification performance. Considering that deeper models generally incur higher computational costs and increased inference time, the 6-layer setup offers a better trade-off between effectiveness and efficiency. Therefore, it can be regarded as the more practical and balanced design choice under constrained resources.
Figure~\ref{fig6}(b) demonstrates that increasing the number of input frames consistently improves recognition accuracy across datasets. This trend suggests that expanding the temporal receptive field allows the model to capture longer-range temporal dependencies and richer motion patterns. Such enhancement is especially beneficial for identifying complex or subtle actions, as more comprehensive temporal cues strengthen the model's ability to discover causal relationships of atomic actions in video.

\section{Conclusion}
In this paper, we propose a \textbf{C}ross-\textbf{M}odal \textbf{D}ual-\textbf{C}ausal \textbf{L}earning (CMDCL) framework for long-term action recognition that integrates textual and visual cues to expose true causal relationships and mitigate spurious correlations. Our approach leverages textual and visual causal intervention modules to debias text embeddings and deconfound visual features without relying on pre-trimmed videos. Extensive experiments on multiple benchmarks demonstrate that CMDCL consistently outperforms previous methods.

\begin{acks}
	This work was supported by the National Natural Science Foundation of China (Nos. 62476015, 62171298) and the Joint Fund of the National Natural Science Foundation of China (No. U21B2038).
\end{acks}

\bibliographystyle{ACM-Reference-Format}
\bibliography{mm25}


\end{document}